%% file: arxiv.tex
\documentclass[runningheads]{llncs}
\usepackage{graphicx}

\usepackage{tikz}
\usepackage{comment}
\usepackage{amsmath,amssymb} 
\usepackage{color}
\usepackage{graphicx}
\usepackage{amsmath}
\usepackage{amssymb}
\usepackage{booktabs}
\usepackage{url}
\usepackage{amsfonts}       
\usepackage{algorithm}
\usepackage{algpseudocode}
\usepackage{multirow}
\usepackage{multicol}
\usepackage{booktabs}
\usepackage{capt-of}
\usepackage[normalem]{ulem}
\usepackage{wrapfig}
\usepackage{glossaries}

\include{math}

\usepackage[pagebackref,breaklinks,colorlinks]{hyperref}

\usepackage[capitalize]{cleveref}
\crefname{section}{Sec.}{Secs.}
\Crefname{table}{Table}{Tables}
\crefname{table}{Tab.}{Tabs.}
\crefname{figure}{Figure}{Figures}
\crefname{algorithm}{Algorithm}{Algorithms}
\crefname{appendix}{Appendix}{Appendices}
\crefformat{equation}{(#2#1#3)}

\newacronym{VAE}{VAE}{variational autoencoder}

\usepackage[accsupp]{axessibility}  

\usepackage[width=122mm,left=12mm,paperwidth=146mm,height=193mm,top=12mm,paperheight=217mm]{geometry}

\begin{document}
\pagestyle{headings}
\mainmatter
\def\ECCVSubNumber{4116}  

\title{Sample-dependent Adaptive Temperature Scaling for Improved Calibration}

\titlerunning{Adaptive Temperature Scaling}
%
\author{Tom Joy\inst{1} \and
Francesco Pinto\inst{1} \and
Ser-Nam Lim\inst{2} \and
Philip H. S. Torr \inst{1} \and \\
Puneet K. Dokania \inst{1,3} \\
\email{\{tomjoy\}@robots.ox.ac.uk}}
\authorrunning{T. Joy et al.}
%
\institute{$^1$University of Oxford, $^2$Meta AI, $^3$Five AI Ltd., UK}

%
\maketitle

\begin{abstract}
It is now well known that neural networks can be wrong with high confidence in their predictions, leading to poor calibration.
The most common post-hoc approach to compensate for this is to perform temperature scaling, which adjusts the confidences of the predictions on any input by scaling the logits by a fixed value.
Whilst this approach typically improves the average calibration across the \emph{whole} test dataset, this improvement typically reduces the individual confidences of the predictions irrespective of whether the classification of a given input is correct or incorrect. 
With this insight, we base our method on the observation that different samples contribute to the calibration error by varying amounts, with some needing to increase their confidence and others needing to decrease it.
Therefore, for each input, we propose to predict a different temperature value, allowing us to adjust the mismatch between confidence and accuracy at a finer granularity.
%
%
%
Our method is applied post-hoc, consequently using very little computation time and with a negligible memory footprint and is applied to off-the-shelf pre-trained classifiers.
We test our method on the ResNet50 and WideResNet28-10 architectures using the CIFAR10/100 and Tiny-ImageNet datasets, showing that producing per-data-point temperatures is beneficial also for the expected calibration error across the whole test set.\footnote{This work is still in progress. \\Code is available at: \href{https://github.com/thwjoy/adats}{https://github.com/thwjoy/adats}}
\end{abstract}

\section{Introduction}

For neural networks to be employed in real-world safety-critical applications, we do not only require them to produce correct predictions, but also provide reliable confidence estimates in their predictions (i.e. they are calibrated). 
Limiting our scope to neural classifiers, using the maximum probability of the predictive distribution as a confidence measure, literature has established that a mismatch exists between such notion of confidence and the expected accuracy. 
Indeed, such models generally suffer from being \emph{on average} overconfident over the test-set.

A simple approach to rectify this issue is to perform \emph{temperature scaling}~\cite{guo2017calibration}, a post-hoc method which scales the logits by a single scalar value, obtained through cross validation.
This approach improves the classifier's performance on standard calibration metrics across a test dataset.
However, from a per-sample point of view there are significant issues.
Since the temperature is found by minimising the calibration error (in expectation) over the \emph{entire} validation set, and since neural networks are overconfident on average, practically speaking, the effect of temperature scaling is to reduce the confidence for every prediction. 
%
However, as we will discuss, different samples contribute by varying amounts to the calibration error.

\begin{figure}[h!]
    \centering
         \includegraphics[width=0.5\linewidth]{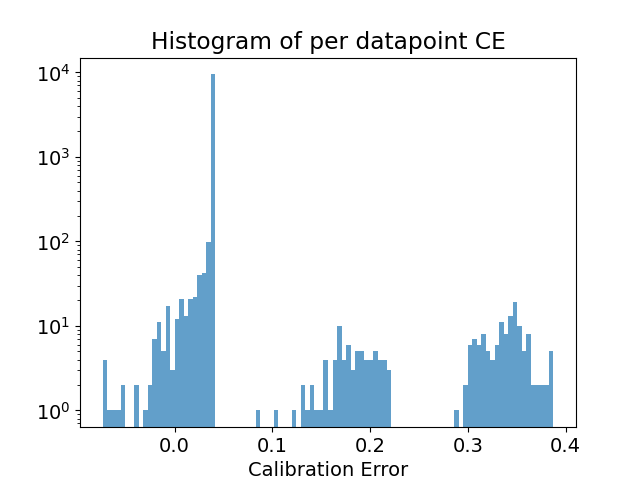}
   
    \caption{Histogram of per sample contribution to calibration error, positive numbers indicate overconfidence. Here we can see that the samples contribute by different amounts to the overall calibration error. Predictions are for CIFAR-10 using a ResNet-50.}\label{fig:error_hist_nn}
    \vspace{-3ex}
\end{figure}

This issue can be seen in \cref{fig:error_hist_nn}, which shows the histogram of the individual contributions to the calibration errors;
i.e. the distribution of $p(\by|\bp_i) - \bp_i$, where $p(\by|\bp_i)$ is the accuracy and $\bp_i$ is the softmax probability for the data point $i$, the calibration error can be obtained by taking the weighted average over all the values.\footnote{Here $p(\by|\bp_i)$ is obtained through histogram binning and represents the accuracy of each bin, and the weights are proportional to the number of samples in each bin.}
Here the mismatch between per data-point confidence and accuracy is not constant across all the data-points, and hence miscalibration cannot be fixed by scaling the logits by a single fixed value, a key assumption in vanilla temperature scaling. 
The calibration error varies significantly, with a small (but not insignificant) number of samples on which the network is overconfident.
Consequently, scaling the predictions with a single temperature value will adjust \emph{all} of the errors in the same way.
Typically, the temperature values obtained are greater than 1, resulting in a reduction of confidence of \emph{all} predictions, regardless of whether they are correct with low confidence or incorrect with high confidence.

To combat this, we propose a method which produces per-data-point predictions of the temperature, permitting an adequate decrease in the confidence on samples which the classifier is \emph{likely} to get wrong, and an increase in the confidence on predictions it is \emph{likely} to get correct. 
As a result, we obtain better test Expected Calibration Error (ECE)\cite{guo2017calibration} both on in-distribution sets (i.e. the test set is i.i.d. with respect to the training set) and under covariate-shifted sets (i.e. the test set shares the same set of labels of the training set, but the inputs are not i.i.d. with respect to the training set). 

Like temperature scaling,  our method is applied post-hoc and is very fast to train and test.
We extensively test the calibration of ResNet50~\cite{he2016deep} and WideResNet28~\cite{zagoruyko2016wide} when using our method on CIFAR10/CIFAR100 and TinyImageNet, including results under data-shift~\cite{hendrycks2019benchmarking}.
Specifically our contributions is to identify a limitation in using a constant temperature for temperature scaling and propose a novel method to predict temperature values on a per-data-point basis to address this limitation.
Our method produces a temperature value that is sample dependent, allowing the method to reduce the confidence of incorrect predictions, but also increase the confidence of correct ones.

\section{Problem formulation}
\subsection{Network Overconfidence and Temperature Scaling}
Given an input $\bx$, a standard $K$-class neural classifier first extracts a feature embedding $\phix$ before computing the logits $\bs = f(\phix) \in \bbR^K$ and finally applying the softmax operator $\bp = \sigma(\bs) = \frac{\exp{(\bs)}}{\sum_i \exp{(s_i)}}$ to obtain the class probabilities for the categorical distribution, the prediction is then given as $\hat{y} = \argmax \bp$. 
%
A classifier is said to be calibrated if the confidence in its prediction (usually taken to be $\max \bp$) matches its accuracy on expectation, i.e. if a classifier makes predictions with a confidence of 80\% for a certain set of points, then it also has an accuracy of 80\% on such set of points.
Typically, the predictions of neural networks are overconfident, i.e. the probability of the predicted class is higher than their expected accuracy~\cite{guo2017calibration}. 

\begin{figure}[h!]
    \centering
    \scalebox{0.9}{
    \begin{tabular}{ccc}
         \includegraphics[width=0.3\linewidth]{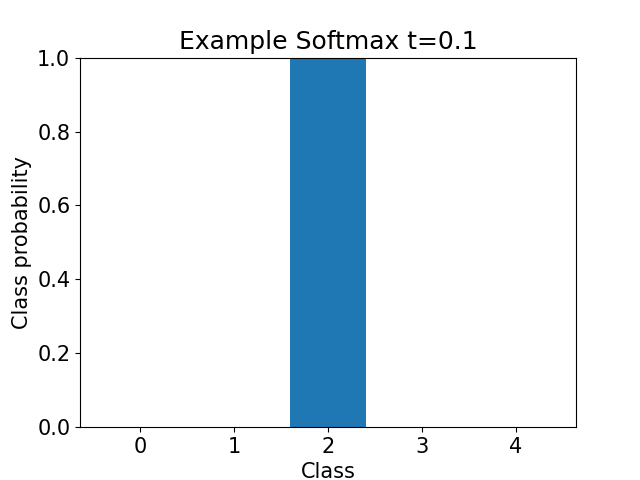}&
         \includegraphics[width=0.3\linewidth]{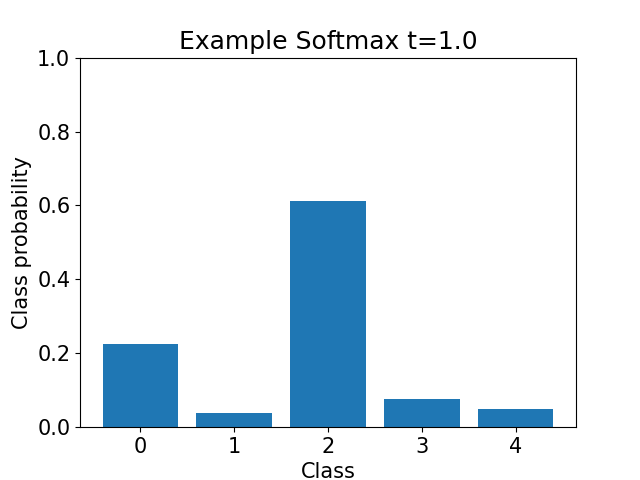}&
         \includegraphics[width=0.3\linewidth]{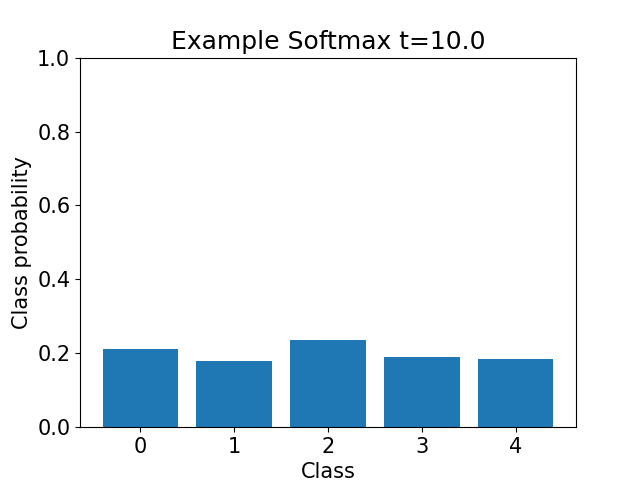}
         
    \end{tabular}
    }
    \caption{Example plots of Softmax distribution with different temperature values for fixed logits. Left to right: $T=0.1, T=1.0$ and $T = 10.0$.}
    \label{fig:softmax}
\end{figure}

Temperature scaling~\cite{guo2017calibration} consists of re-scaling the logits by a constant factor $T \in \bbR^+$ before applying the softmax, i.e. $\bp' = \sigma(\frac{\bs}{T})$.
The value of $T$ can drastically affect the entropy of the predicted distribution, which is demonstrated in~\cref{fig:softmax}, where a value of $T > 1$ leads to a higher entropy distribution (the higher $T$, the higher the entropy); a value of $T < 1$ leads to a lower entropy distribution (the lower $T$, the more ``peaky'' the distribution).
%

The temperature $T$ is usually found by minimising the ECE or the Negative Log-Likelihood (NLL) using a validation set. Typical optimal values for $T$ are usually greater than $1$~\cite{mukhoti2020calibrating}, indicating that, on average, optimising the ECE or NLL across the validation set leads to a higher entropy of the predictions. 
However, this approach decreases the confidences of \emph{all} the predictions without considering that the miscalibration error can  vary widely on a data-point basis.
For correct predictions, temperature scaling will make the predictions more under-confident, whilst for incorrect predictions, the temperature may not be the right value to bring the confidences down to a level which will make it calibrated.

Loosely speaking, this suggests that further improvements in calibration can be achieved by using a variable temperature $T$, predicted on a per data-point basis (i.e. $T=g(\bx)$), permitting $T > 1$ for samples which are correctly classified, and $T < 1$ for the incorrect ones.
This enables the model to have a greater flexibility when compensating for miscalibration, as it respects the individual contributions each sample makes to the ECE.
%
%
Moreover, this approach can be applied \emph{without} affecting a classifiers accuracy\footnote{For a proof please see \cref{app:temp}}.
%

\subsection{Why Jointly Learning Temperature Alongside the Network Weights Might Go Wrong?}\label{sec:gradients}

One might suggest introducing the temperature as one of the outputs of the network and jointly learning it as part of the training process. Here we outline why this approach of learning to predict the temperature values $T$ and the predictive probabilities $\bp$ might result in suboptimal performance.
Consider the last layer of a NN with parameters $\bw \in \bbR^{D \times K}$ for a feature space of size $D$ and the cross entropy loss $\calL : \bbR^K \rightarrow \bbR$.
The gradient for the layer is given as
\begin{align}
    \frac{\partial\calL}{\partial\bar{\bw}} = \frac{\partial\bs}{\partial\bar{\bw}}(\sigma(\bs) - \bq),
\end{align}
where $\bq \in \Delta^K$ denotes the one-hot ground-truth label and $\sigma(\bs) - \bq = \{\sigma(s_j) - q_j : j \in \{1\dots K\}$\footnote{See \cref{app:gradients} for a derivation}, $\bar{\bw}$ indicates the network weights are flattened to column vector form.
Inspecting the gradients indicates that the gradient starts to vanish when $s_k \rightarrow \infty$ and $s_{\text{\textbackslash} k} \rightarrow -\infty$, where $k$ is the correct class.
Or to put it simply, the optimisation does not converge until the network produces one-hot logits.

This forces the magnification of the network weights ~\cite{mukhoti2020calibrating}, which subsequently leads to an overconfident network and hence miscalibrated predictions.
A mechanism to achieve the desired one-hot prediction without magnifying the weights could be instead to na\"ively learn the temperature alongside the logits, assuming the model is trainable and converges.
In this case, gradient updates would decrease the value of $T$, resulting in a lower-entropy distribution that is more ``peaky''.
%
%
We now discuss why this approach might not work well in practice.

If we consider the gradient of the temperature, which is given as
\begin{align}
    \frac{\partial\calL}{\partial T} = \sum_k \frac{q_k}{T^2}\bigg(s_k\sum_{i\text{\textbackslash}k}\exp\Big(\frac{s_i}{T}\Big) - \sum_{j\text{\textbackslash}k}s_j\exp\Big(\frac{s_j}{T}\Big)\bigg),
\end{align}
which decreases the value of $T$ for a correct prediction ($\tilde{k} = \argmax_{k}s_k)$, leading to more confident predictions, see \cref{app:temp_grad} for a proof.
Typically, the train accuracy will approach 100\%, meaning that gradient updates to $T$ cause it to decrease without any moderation, preventing the network from learning how to predict $T$ appropriately.
In short, there is essentially only data for correct predictions, \textit{preventing crucial information on how the network should behave when it's wrong}.
Consequently, learning $T$ na\"ively is not a feasible option as the network just learns to be confident everywhere.
%

\subsection{Learning to Calibrate}\label{sec:cal_learning}
For the reasons discussed above, we  propose an approach that, similar to the standard temperature scaling of~\cite{guo2017calibration}, works on a already trained model. However, as opposed to~\cite{guo2017calibration}, our approach involves a new temperature prediction module (a small neural network) that operates on each input sample independently and whose objective is to extract information from the trained model itself in order to calibrate the confidences of each prediction. We call this learning to calibrate.

%
Doing so requires learning a temperature prediction module on a data-set consisting of data-points $\calX_{cal} = \{\bx_n\}^N, \calX_{train} \cap \calX_{cal} = \O$, neural network predictions $\calP_{cal} = \{\bp_n\}^N$ and labels $\calY_{cal} = \{\by_n\}^N$.
It is important to note that the objective here is to learn to assign low confidences to data points which are \emph{likely} to be incorrect and high confidences to those which are likely to be correct.

Specifically for a given data-point $\bx \in \calX_{cal}$, we propose to  optimise the temperature prediction module over $T$ by maximising the log probability of the label $\by$ under the Categorical probability distribution parametrised by the $T$-scaled logits $\bs$, i.e. $T^* = \argmax_T\log\text{Cat}(\by ; \softmax{\bs / T})$\footnote{Which is equivalent to the cross entropy loss.}.
Here we do not optmise $\bs$ but keep it fixed;  we are only optimising w.r.t to $T$.

In situations where $\by=\argmax_k\bp_k$ (i.e. correct prediction), the target function is maximised when $T \rightarrow 0$, as we want the predicted probabilities to match the one-hot logits, e.g. see $T=0.1$ in \cref{fig:softmax}.
This is equivalent to minimising the entropy of the predictive distribution by only manipulating $T$, which is the desired outcome for a correct prediction.

In situations where the prediction is incorrect, $\by \neq \argmax_k\bp_k$, to maximise the target function we need to maximise $\bp_{\by}$ and minimise $\bp_{\tilde{k}}$, where $\tilde{k} = \argmax_k\bp_k$.
As the temperature prediction module cannot change the predicted label, the optimzation accepts the incorrect prediction and maximise the target function by flattening the Softmax outputs with $T >> 1$, which is equivalent to maximising the entropy of the predictive distribution.
%
%
This effect can be seen by considering the case where predicting class 2 in~\cref{fig:softmax} is the incorrect prediction; among the three cases shown, $T = 10$ maximises the $\text{Cat}(\by \neq 2; \softmax{\bs / T})$.

\section{Adaptive Temperature Scaling}\label{sec:method}
We are now ready to outline the specifics of our proposed temperature prediction module. 
Given a data-set $\calX_{cal}$, we want to learn which samples the classifier should be confident about and which it should not.
Rather than acting on the image space, we instead use the feature extractor of the classifier, as it has already learnt how to extract the information needed for class prediction and also contains a notion of the associated confidence. Additionally, working only on the feature space, as done in ~\cite{guo2017calibration}, has already provided highly promising results in calibrating models for a variety of tasks. Therefore, assuming that the feature space already contains good amount of information in order to learn to calibrate the confidences is reasonable. 
What we now need is a method to extract this confidence information from the feature space and leverage it appropriately to calibrate the predictions.

\subsubsection{Representing Uncertainty with the \gls{VAE}}
\glspl{VAE}~\cite{kingma2013auto} act as an efficient model to obtain representations of data; the representations encapsulate the generative factors in a lower-dimensional subspace and are rich enough to reconstruct the data sample.
In the specification of the generative model, the user has to specify a prior over the latent variables (typically an isotropic Gaussian) where the KL distance between the prior and approximate posterior is minimised during training.
Unlike a standard autoencoder~\cite{hinton1994autoencoders}, there is now a mechanism to obtain a likelihood on the latent codes.
In reality this value forms part of the importance weight and can be used as a proxy to the true likelihood but avoids issues associated with deep generative models~\cite{nalisnick2018do}.

From a mechanistic point of view, we expect samples which are much more common to be placed in the centre of the prior.
Here we leverage this idea and use the latent likelihoods as a basis 
to predict the temperature value.
Indeed, we find empirically that this approach works well in practice.
Rather than using a standard \gls{VAE} which places an isotropic Guassian as the prior we instead introduce a prior for each class, which is parameterised by $\{\lambda_k\}^K$, allowing the classes to cluster individually.
This prevents any issues with clusters for individual classes being placed in lower likelihood regions of the latent space.
With this mixture prior, the evidence lower bound is given as
\begin{align}
\log p(\phix) \ge \elbo{\phix} = \bbE_{q_\varphi(\bz|\phix)}\log\frac{p_\vartheta(\phix|\bz)
\pzy}{q_\varphi(\bz|\phix)},
\end{align}
where $q_\varphi(\bz|\phix), p_\vartheta(\phix|\bz)$ and $\pzy$ represent the encoder, decoder and conditional prior, the parameters of the VAE are given as $\mathbf{\Theta} = \{\vartheta, \varphi, \lambda_1, \dots, \lambda_K\}$.
The parameters of the conditional prior are learnt alongside the parameters of the encoder and decoder\cite{tomczak2018vae}, each component of the prior is modelled using a Gaussian, i.e. $\boldsymbol{\lambda}_k = \{\boldsymbol{\mu}_k, \boldsymbol{\sigma}_k\} \in \bbR^{D_\bz}$, where $D_\bz$ is the size of the latent space.
The use of this conditional prior forces the aggregate posterior for \emph{each} class to match a Gaussian distribution, i.e. $\sum_{\bx\in\calX_k}q(\bz|\phix) \approx \calN(\bz; \boldsymbol{\mu}_k, \boldsymbol{\sigma}_k)$.
This encourages the representations for each class to cluster around a known distribution $\pzy$, which we will use to obtain a pseudo likelihood to predict the temperature value. 
The choice of the VAE was in part down to the motivation that hard samples will have lower latent-likelihood but also because empirically we found it worked well in practice.
Before outlining the details of the approach in in~\cref{sec:temp_net} and ~\cref{sec:cal_learn}, we first provide evidence of this empirical motivation to use a \gls{VAE}.

\begin{figure}
    \centering
    \begin{tabular}{cc}
         \includegraphics[width=0.45\linewidth]{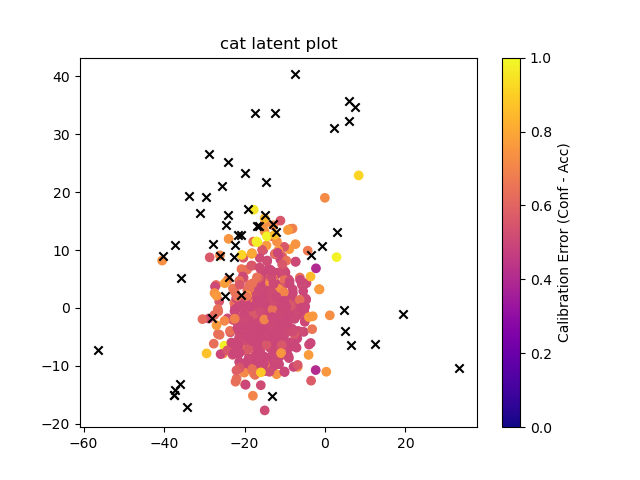}
         &
         \includegraphics[width=0.45\linewidth]{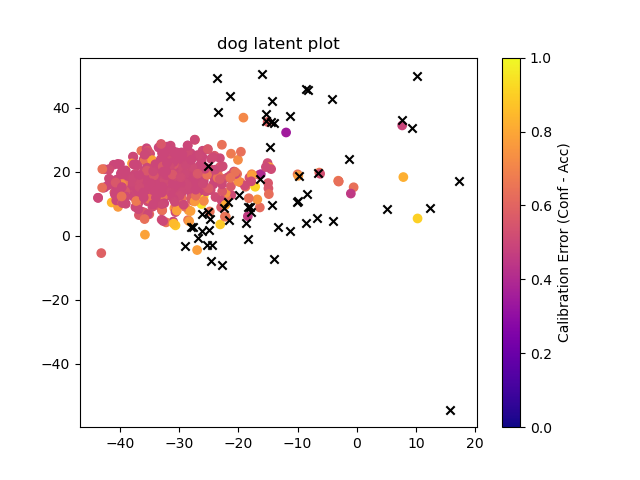}
    \end{tabular}
    \caption{t-SNE plot for classes \texttt{cat} and \texttt{dog}, colour indicates per data-point contribution to ECE, 0.5 indicates no contribution to ECE. Generally, samples with little contribution to calibration error (pink) are placed around the centre of the cluster, unlike samples with a high contribution (yellow and orange) which are placed near the edges. Furthermore, incorrect samples (black cross) are placed significantly far away from the cluster centre.}
    \label{fig:class_tsne}
\end{figure}

We now perform a preliminary experiment, which serves to investigate which samples in the latent representation contribute the most to the calibration error.
Specifically, we construct a t-SNE plot for each class of CIFAR-10 but colour code the points depending on their per-data-point contributions to the calibration error.
This provides a visual method for us to inspect where samples which harm calibration are placed, which can be seen in \cref{fig:class_tsne}.
Here we can see that data-points which do not contribute to the calibration error tend to be placed near the centre of the cluster, and ones which do, or are incorrect, indicated by a black cross, are placed far from the centre.
This highlights that the \gls{VAE} is at able, to some extent, to provide a basis to predict the temperature, we then utilise this representation to predict $T$ through a simple Multi Layer Perceptron.

%

\subsection{Temperature Prediction Network}\label{sec:temp_net}
\begin{figure}[h]
    \vspace{-5.0ex}
    \centering
    \includegraphics[width=0.5\linewidth]{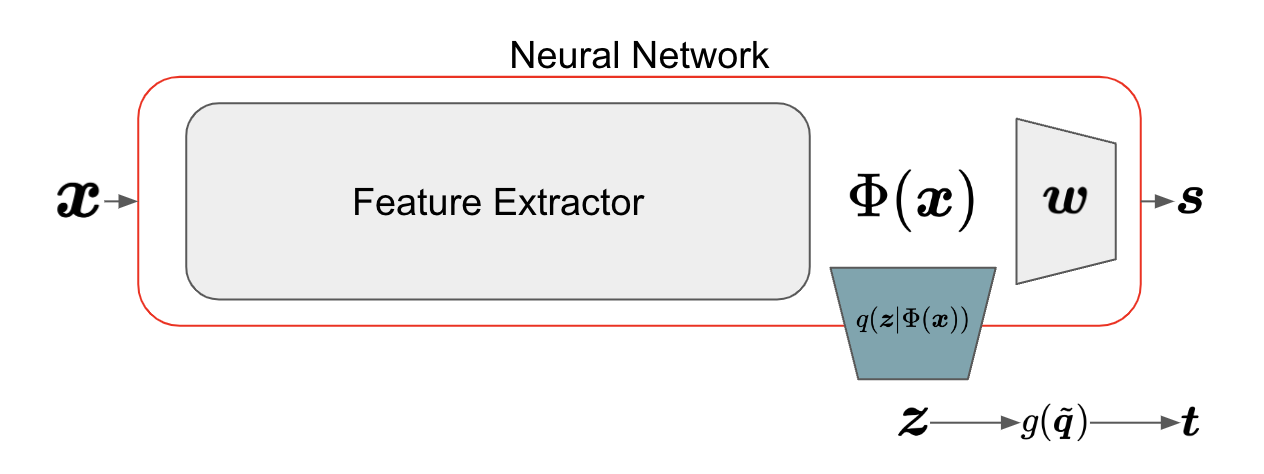}
    \vspace{-3.0ex}
	\caption{High level architecture. The off shelf neural-network is represented by the red box, where the parameters are left unchanged, the learnable VAE encoder is indicated by $q(\bz|\phix)$, with the $g_\theta(\tilde{\bq})$ as the MLP predicting $T$.}
	\label{fig:graphical_model}
	\vspace{-5.0ex}
\end{figure}

Given that the VAE structures the latent in space in a way which makes it amendable to confidence prediction, we learn a very simple MLP parameterised by $\theta$, which predicts the temperature based on the latent embeddings, using the cross entropy loss as an objective.
%
%
Rather than using the latent samples as input to the MLP, given the observations in \cref{fig:class_tsne}, we choose to predict the temperature as a function of the vector of log-likelihoods on \emph{all} of the conditional priors, specifically $T = g_\theta(\tilde{\bq})$ where $g : \bbR^K \rightarrow \bbR$ is the MLP which predicts the temperature and $\tilde{\bq} = \{\log p_{\lambda_k}(\bz|\by_k)|\forall k\}$, i.e each element $\tilde{\bq}_i$ contains the log-likelihood of $\bz$ on the corresponding conditional prior $p_{\lambda_k}(\bz|\by_k)$.
Evaluating $\log \pzy$ can be viewed as a pseudo likelihood of $\bx$, consequently the module predicts the temperature as a non-linear transform of a pseudo-likelihood of the sample.
It is also important to point out that due to the use of feature space as the input, we are able to use small architectures, making this approach very fast during training and at test time. 
We represent a high level overview and the graphical model in ~\cref{fig:graphical_model}.

\subsection{Calibrated Training Details}\label{sec:cal_learn}
The overall post-hoc learning algorithm is very simple and the module can be trained in under a minute on an 8Gb Titan Xp for most datasets, depending on the validation set and feature space size, we give an overview of the procedure in \cref{alg:train}.
We combine learning the VAE and the temperature prediction network into one objective. 
Specifically, we maximise the following objective
\begin{align}
    \calL(\bx, \by) &= \elbo{\phix} +  \log\text{Cat}(\by \mid \softmax{\bs / g_\theta(\tilde{\bq})}
\end{align}
with $\tilde{\bq} = \{\log \pzy|\forall \by\}\quad\bz \sim \qzx$ and using Normal distributions for $\bz$; Laplace distribution over $\phix$ (L1 loss); and a Categorical over $\by$.
To train the VAE, we use a held-out dataset from training the network, i.e. $\calX_{train} \cap \calX_{cal} = \O$.
We used the Adam optimiser with a learning rate of $0.001$ and trained for 50 epochs.
This is an additional benefit of training the VAE on the $\phix$, as the feature space has a lower dimensionallity and simpler structure than the image space, leading to much faster training and the ability to use simpler networks. 

\begin{algorithm}
\caption{Learning Adaptive Temperature (Ada TS)}\label{alg:train}
\begin{algorithmic}
\Require $\calX_{cal}, \calY_{cal}, \calP_{cal}$
\While{not converged}
    \State $\bx, \by \gets$ Random batch
    \State $\nabla_{VAE} \gets \nabla \elbo{\phix}$ \Comment{Standard VAE gradient}
    \State $\tilde{\bq} = \{\log \pzy|\forall \by\}\quad\bz \sim q(\bz|\phix)$ \Comment{Vector of pseudo likelihoods}
    \State $\nabla_{T} \gets \nabla\log\text{Cat}(\by; \softmax{\bs / g_\theta(\tilde{\bq}})$ \Comment{CE loss for temperature}
    \State $\{\boldsymbol{\Theta}, \theta\}_{t+1} \gets \{\boldsymbol{\Theta}, \theta\}_t - \alpha(\nabla_{VAE} + \nabla_{T})$ \Comment{Update parameters $\{\boldsymbol{\Theta}, \theta\}$}
\EndWhile
\end{algorithmic}
\end{algorithm}
\vspace{-4ex}

\paragraph{Calibration at Test Time}
During test time, the features of the data point $\phix$ and predicted logits are computed from the classifier $\bs = f(\phix)$, the temperature can then predicted through $T = g_\theta(\tilde{\bq})$.
The calibrated predictions are then computed as $\bp = \sigma(\bs / T)$.

%

\section{Related Work}
\paragraph{Uncertainty Estimation}
In deep learning, the most typical way to address uncertainty estimation is to make the networks output a distribution, and to extract an uncertainty measure as a function of the predictive distribution. 
Bayesian approaches define a prior distribution over the weights of the network and apply inference techniques to update such distributions given the training set. Given the intractability of exact inference for neural networks, several approximate variational inference schemes have been proposed~\cite{gal2016dropout,BayesByBackprop,BayesDropout1,SGLD,cSGLD,sotchGradHMC}. 
Recent literature tries to combine the benefits of Bayesian deep learning with the training of deterministic neural networks trained via standard optimisation algorithms. 
Some methodologies suggest using a Laplace approximation of a trained network to approximate a Gaussian  using a Laplace approximation around the optimal parameters \cite{Ritter2018aLaplace,HeinBeingBayesianABit2020}. 
Others suggest replacing the head of the network with a Gaussian Process \cite{liu2020simple} or a head parametrising a Dirichlet distribution \cite{malinin2018predictive,BeingBayesianCategorical}, or just performing Bayesian inference on the final layer~\cite{riquelme2018deep}.
Another family of models  leverages ensembles~\cite{lakshminarayanan2016simple} to output distributions. Given the extreme computational and memory requirements of ensembles, several techniques have been suggested to obtain the ensembling benefits more efficiently ~\cite{havasi2020training,Wen2020-BatchEnsemble} 

%
%
%
%
%
%

\paragraph{Calibration}
Deep Neural Networks suffer from overconfident classification scores, Which can be alleviated through temperature scaling in post-processing~\cite{guo2017calibration}---a modern variant of Platt scaling~\cite{platt1999probabilistic}.
As previously mentioned, this typically comes at the cost of decreasing the confidence in correct predictions~\cite{kumar2018trainable}.
Other approaches include histogram binning~\cite{zadrozny2001obtaining}; isotonic regression~\cite{zadrozny2002transforming}; Bayesian binning~\cite{naeini2015obtaining,naeini2016binary}; and bin-wise temperature scaling~\cite{ji2019bin}.
Overconfidence is caused by over-fitting to the cross-entropy loss, which can be alleviated by instead using a focal loss ~\cite{lin2017focal,mukhoti2020calibrating}.
In a similar fashion, \cite{kumar2018trainable} utilised a differentiable proxy during training to improve calibration.
Label smoothing was also shown to improve calibration~\cite{muller2019does}.
It has also been shown that recomputing the coefficients of batch normalization improves calibration~\cite{nado2020evaluating}.
Tangentially, \cite{ovadia2019can} performed a large scale comparison of methods under dataset-shift.
A similar method to ours is~\cite{ding2021local}, which tackles the problem of semantic segmentation calibration by predicting per-data-point and per-pixel temperature values.
Another method which is of note is~\cite{kull2019beyond}, which transforms the softmax predictive distribution into a Dirichlet distribution.


\section{Results}
Before evaluating the model, we define the hypothesis we are trying to test.
Specifically, we want to evaluate if predicting the temperature on a per-data-point basis leads to improved calibration over vanilla temperature scaling.
Secondly, we wish to investigate how adaptive temperature performs under dataset shift.

We performed our experiments on WideResNet28-10~\cite{zagoruyko2016wide} and ResNet50~\cite{he2016deep} architectures.
We report calibration results on CIFAR10/CIFAR100~\cite{krizhevsky2009learning} and Tiny-ImageNet~\cite{torralba2008tinyimages}.
We conducted distribution-shit experiments using variants CIFAR10-C/CIFAR100-C to test for domain shift~\cite{hendrycks2019benchmarking}.
We used the following as models for our evaluation:
\begin{itemize}
    \item Cross Entropy Loss, due to it's popularity and wide adoption.
    \item Brier Score~\cite{brier1950verification}, due to it's ability to obtain well calibrated predictions~\cite{mukhoti2020calibrating}.
    \item Deep Ensembles~\cite{lakshminarayanan2016simple}, as it achieves state of the art results.\footnote{Applying adaptive temperature scaling to deep ensembles does not necessarily preseve accuracy, however as we show in our experiments the difference is negligible.}
\end{itemize}
Results are obtained for multiple seeds for Cross Entropy and Brier Score, but only one seed for Deep Ensembles due to the number of models needed.

%

\subsection{Calibration}
Here we evaluate how adaptive temperature scaling affects standard calibration metrics compared to vanilla temperature scaling.
We report results using the ECE, which divides the probability into equally sized bins and then computes the absolute difference between confidence and accuracy for each bin before taking the average.
However, the ECE is known to be a biased estimator of the theoretical probabilistic expectation~\cite{CalibrationRevisitingBiases}, whose performance depends on the binning size and on the distribution of samples in each bin. For this reason, the ECE reliability as a good miscalibration metric is being questioned and several alternatives have been proposed (e.g. ~\cite{MeasuringCalibration,MitigatingBiasCalibration,mukhoti2020calibrating}). Among these, we choose to also use the AdaECE~\cite{mukhoti2020calibrating}, which uses adaptive bin sizes to ensure each bin contains the same number of samples. 

\begin{figure}[h!]
	\centering
	\vspace{-5ex}
	\begin{tabular}{ccc}
		\includegraphics[width=0.3\linewidth]{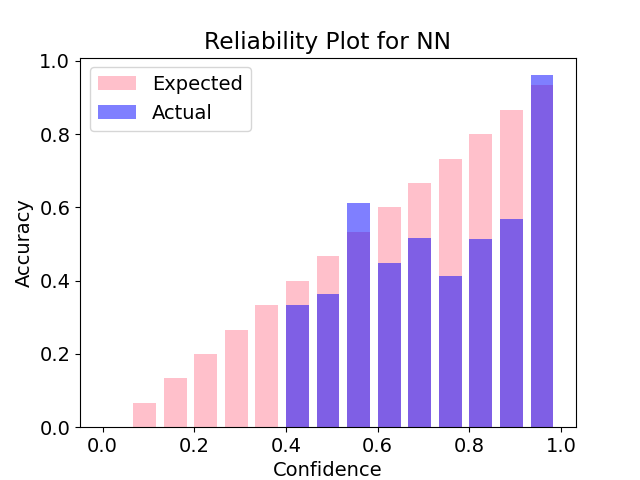} &
		\includegraphics[width=0.3\linewidth]{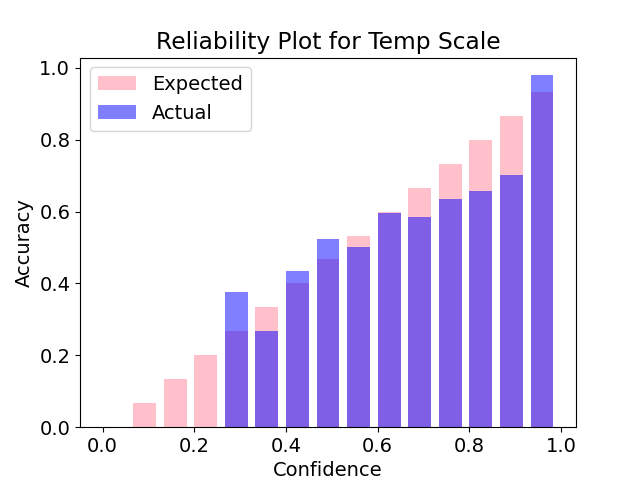} &
		\includegraphics[width=0.3\linewidth]{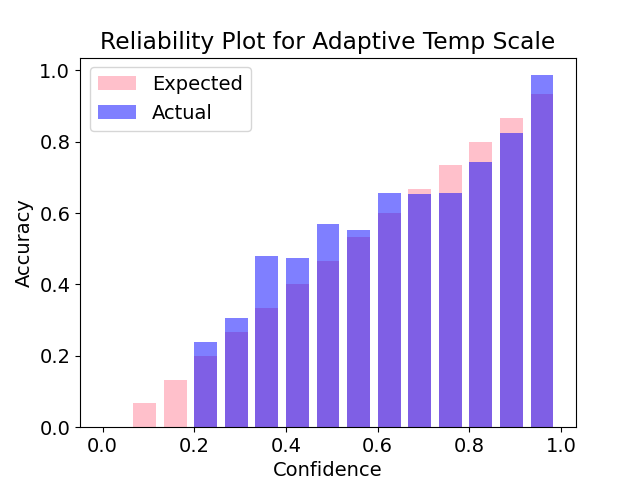}
	\end{tabular}\caption{Reliability plots for: left) vanilla predictions; middle) temperature scaling; right) adaptive temperature scaling (ours). Temperature scaling was optimised through cross validating in the range 0 - 10 and optimised the ECE. CIFAR-10 on ResNet50.}\label{fig:reliability_plots}
\end{figure}

We report the results in \cref{tab:cal} along with reliability plots in \cref{fig:reliability_plots}, where it can be seen that adaptive temperature scaling improves calibration compared to standard temperate scaling.
In all cases our method is able to outperform vanilla temperature scaling, with large improvements obtained when using the cross entropy loss, e.g. $0.93 \rightarrow 0.76$ and $3.76 \rightarrow 2.95$ ECE for CIFAR10 and CIFAR100 when using the WideResNet2810 Network.

\begin{table}[h]
    \centering 
\scalebox{0.7}{
	\begin{tabular}{cc|ccc|ccc}
		\toprule
		\midrule
		\textbf{Method} & \textbf{Scaling} & \textbf{Accuracy} ($\uparrow$) &  \textbf{ECE} ($\downarrow$) & \textbf{AdaECE} ($\downarrow$) & \textbf{Accuracy} ($\uparrow$) &  \textbf{ECE} ($\downarrow$) & \textbf{AdaECE} ($\downarrow$)\\
            \cmidrule(lr){1-8}
            &&\multicolumn{6}{c}{\textbf{CIFAR10}}\\
            \midrule
            &&\multicolumn{3}{c}{\textbf{WideResNet2810}}&\multicolumn{3}{c}{\textbf{ResNet50}}\\
            \cmidrule(lr){1-8}
            CE&None & 95.52 $\pm$ 0.43  & 2.15 $\pm$ 0.18  & 2.13 $\pm$ 0.18 & 93.13 $\pm$ 1.97  & 3.75 $\pm$ 1.32  & 3.74 $\pm$ 1.32 \\ 
            CE&Vanilla TS & 95.52 $\pm$ 0.43  & 0.93 $\pm$ 0.20  & 0.98 $\pm$ 0.30 & 93.13 $\pm$ 1.97  & \textbf{1.41 $\pm$ 0.43}  & \textbf{1.45 $\pm$ 0.44} \\
            CE&Adaptive TS & 95.52 $\pm$ 0.43  & \textbf{0.76 $\pm$ 0.07}  & \textbf{0.86 $\pm$ 0.20} & 93.13 $\pm$ 1.97  & \textbf{1.13 $\pm$ 0.60}  & \textbf{1.09 $\pm$ 0.57}\\
            \cmidrule(lr){1-8}
            Brier&None & 95.84 $\pm$ 0.10  & 0.92 $\pm$ 0.13  & 1.50 $\pm$ 0.16 & 94.59 $\pm$ 0.23  & 2.03 $\pm$ 0.13  & 2.27 $\pm$ 0.12 \\ 
            Brier&Vanilla TS & 95.84 $\pm$ 0.10  & 1.88 $\pm$ 0.23  & 1.94 $\pm$ 0.19 & 94.59 $\pm$ 0.23  & \textbf{1.67 $\pm$ 0.24}  & 2.08 $\pm$ 0.30 \\
            Brier&Adaptive TS & 95.84 $\pm$ 0.10  & \textbf{1.65 $\pm$ 0.15}  & \textbf{1.61 $\pm$ 0.13} & 94.59 $\pm$ 0.23  & \textbf{1.61 $\pm$ 0.40}  & \textbf{1.53 $\pm$ 0.44}\\
            \cmidrule(lr){1-8}
            Ensmbls&None & 96.35 & 1.68 & 1.61 & 95.62 & 1.92 & 1.89 \\
            Ensmbls&Vanilla TS & 96.35  & 0.61 & 0.68 & 95.62  & 0.93  & 0.84 \\
            Ensmbls&Adaptive TS & 96.37  & \textbf{0.51}  & \textbf{0.46} & 95.64  & \textbf{0.60}  & \textbf{0.58} \\
            \cmidrule(lr){1-8}
            &&\multicolumn{6}{c}{\textbf{CIFAR100}}\\
            \midrule
            &&\multicolumn{3}{c}{\textbf{WideResNet2810}}&\multicolumn{3}{c}{\textbf{ResNet50}}\\
            \cmidrule(lr){1-8}
            CE&None & 80.71 $\pm$ 0.17  & 5.76 $\pm$ 0.16  & 5.70 $\pm$ 0.16 & 77.91 $\pm$ 0.33  & 9.39 $\pm$ 0.42  & 9.37 $\pm$ 0.43 \\ 
            CE&Vanilla TS & 80.71 $\pm$ 0.17  & 3.76 $\pm$ 0.29  & 3.68 $\pm$ 0.28 & 77.91 $\pm$ 0.33  & \textbf{3.63 $\pm$ 0.21}  & \textbf{3.61 $\pm$ 0.26} \\
            CE&Adaptive TS & 80.71 $\pm$ 0.17  & \textbf{2.95 $\pm$ 0.41}  & \textbf{2.90 $\pm$ 0.47} & 77.99 $\pm$ 0.33 & \textbf{3.30 $\pm$ 0.50}  & \textbf{3.32 $\pm$ 0.47}\\
            \cmidrule(lr){1-8}
            Brier&None & 79.25 $\pm$ 0.14  & 4.19 $\pm$ 0.24  & 4.13 $\pm$ 0.22 & 76.03 $\pm$ 0.55  & 4.15 $\pm$ 0.22  & 4.04 $\pm$ 0.25 \\ 
            Brier&Vanilla TS & 79.25 $\pm$ 0.14  & \textbf{3.87 $\pm$ 0.62}  & \textbf{3.90 $\pm$ 0.62} & 76.03 $\pm$ 0.55  & \textbf{3.34 $\pm$ 0.46}  & \textbf{3.41 $\pm$ 0.39 }\\
            Brier&Adaptive TS & 79.25 $\pm$ 0.14  & \textbf{3.67 $\pm$ 0.82}  & \textbf{3.64 $\pm$ 0.74} & 76.03 $\pm$ 0.55 & \textbf{3.30 $\pm$ 0.50}  & \textbf{3.32 $\pm$ 0.47}\\
            \cmidrule(lr){1-8}
            Ensmbls&None & 83.19 & 4.24 & 4.21 & 80.90 & 6.59 & 6.29  \\
            Ensmbls&Vanilla TS & 83.18 & 3.71 & 3.55 & 80.90 & 3.22 & 3.16 \\
            Ensmbls&Adaptive TS & 83.22 & \textbf{2.95}  & \textbf{2.66} & 80.86 & \textbf{2.79}  & \textbf{2.77} \\
            \cmidrule(lr){1-8}
            &&\multicolumn{6}{c}{\textbf{Tiny-ImageNet}}\\
            \midrule
            &&\multicolumn{3}{c}{\textbf{WideResNet2810}}&\multicolumn{3}{c}{\textbf{ResNet50}}\\
            \cmidrule(lr){1-8}
            CE&None & 60.47 $\pm$ 0.17  & 7.54 $\pm$ 4.00  & 7.53 $\pm$ 4.09 & 55.27 $\pm$ 2.19  & 8.92 $\pm$ 2.72  & 8.93 $\pm$ 2.74 \\ 
            CE&Vanilla TS & 60.47 $\pm$ 1.06  & 6.28 $\pm$ 2.43  & 6.15 $\pm$ 2.47 & 55.27 $\pm$ 2.19  & 7.64 $\pm$ 1.53  & 7.57 $\pm$ 1.58 \\ 
            CE&Adaptive TS & 60.04 $\pm$ 1.20  & \textbf{5.18 $\pm$ 1.40}  & \textbf{5.17 $\pm$ 1.32} & 55.27 $\pm$ 2.19  & \textbf{4.51 $\pm$ 1.76}  & \textbf{4.45 $\pm$ 1.78}\\
            \cmidrule(lr){1-8}
            Brier&None & 50.23 $\pm$ 0.45  & 5.56 $\pm$ 0.63  & 5.52 $\pm$ 0.64 & 42.38 $\pm$ 1.21 & 5.33 $\pm$ 1.47  & 5.37 $\pm$ 1.47 \\ 
            Brier&Vanilla TS & 50.23 $\pm$ 0.45 & \textbf{4.55 $\pm$ 0.28} & \textbf{4.43 $\pm$ 0.63} & 42.38 $\pm$ 1.21 & 3.08 $\pm$ 0.59  & 3.12 $\pm$ 0.55 \\
            Brier&Adaptive TS & 50.23 $\pm$ 0.45 & \textbf{4.43 $\pm$ 0.47}  & \textbf{4.21 $\pm$ 0.51} & 42.38 $\pm$ 1.21  & \textbf{2.71 $\pm$ 0.08}  & \textbf{2.60 $\pm$ 0.23}\\
            \cmidrule(lr){1-8}
            Ensmbls&None & 66.16 & 6.21 & 6.19 & 61.90 & 8.89 & 9.00 \\
            Ensmbls&Vanilla TS & 66.16  & 5.12  & 5.06 & 61.90  & 4.29  & 4.43 \\
            Ensmbls&Adaptive TS & 66.00  & \textbf{4.58}  & \textbf{4.41} & 61.76  & \textbf{4.26}  & \textbf{4.17}  \\
            \cmidrule(lr){1-8}
            \midrule 
            \bottomrule
            	\end{tabular}%
            }
    \caption{Calibration results, here we can see that adaptive temperate scaling is able to improve calibration on a variety of models. Bold indicates best results, or with in one standard deviaiton of best results.}
    \vspace{-7ex}
    \label{tab:cal}
\end{table}

\begin{figure}[h]
    \centering
    \begin{tabular}{cc}
         \includegraphics[width=0.4\linewidth]{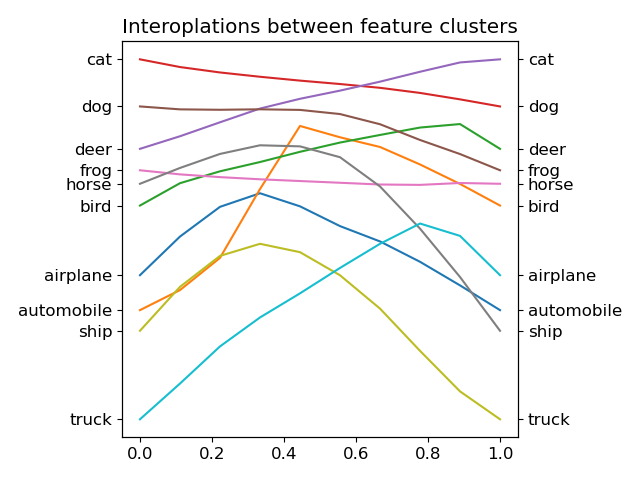} & 
         \includegraphics[width=0.4\linewidth]{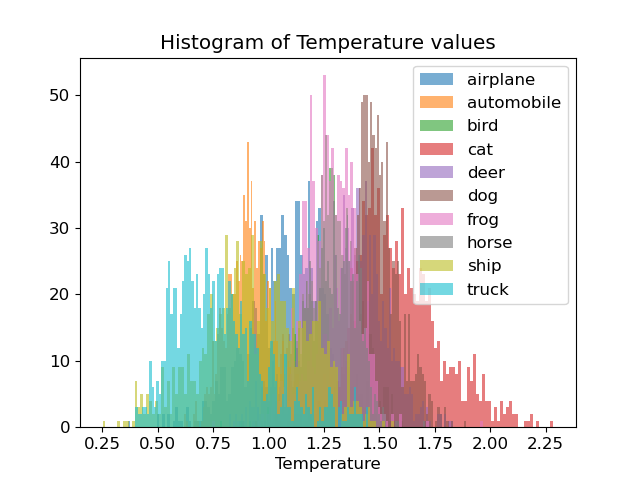}
    \end{tabular}
    \caption{Left: How temperature varies when interpolating between class feature means. Here we can see that temperature increases between classes or remains high for classes who's embeddings are close together. Pairs were chosen to improve visual clarity. Dataset: CIFAR-10; architecture: ResNet50. Right: Histogram of temperature values for each image in CIFAR-10, here we can see that typically objects have a lower temperature than animals, indicating they are easier to classify. Dataset: CIFAR-10; architecture: ResNet50.}
    \label{fig:interps}
\end{figure}

\subsubsection{Data-Shift}
\begin{wrapfigure}{r}{0.45\textwidth}
    \centering
    \begin{tabular}{cc}
        \includegraphics[width=0.9\linewidth]{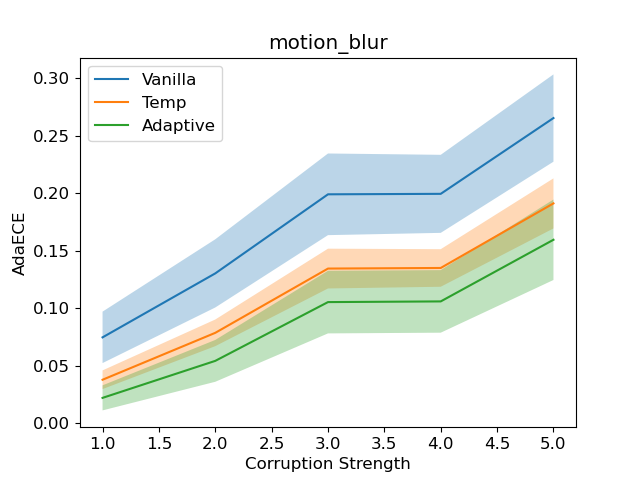} 
    \end{tabular}
    \caption{How AdaECE changes with varying levels of motion-blur corruptions. Adaptive temperature consistently produces lower error rates. CIFAR10-C on ResNet50.}
    \vspace{-6.0ex}
    \label{fig:motion-blur}
\end{wrapfigure}
A key hypothesis we want to test is how adaptive temperature scaling behaves under data-shift.
Specifically, we use the widely used CIFAR10-C and CIFAR100-C datasets, which are corrupted versions of the CIFAR10 and CIFAR100~\cite{hendrycks2019benchmarking}.

The dataset consists of standard CIFAR images which have undergone 15 synthetic corruptions (e.g. noise, weather conditions, image properties) at varying levels.
Within this scenario, the classifier should either be robust to such corruptions (retaining accuracy) or if the accuracy is compromised, reduce the confidences accordingly.
As such, we report the test accuracy as well as ECE and AdaECE in \cref{tab:w2810-corr-cal}, where adaptive temperature scaling shows improvements over temperature scaling.

We also expect to see adaptive temperature scaling provide improvement over temperature scaling as the intensity of corruptions are increased for CIFAR-10-C.
We generate plots highlighting the AdaECE calibration metric as the level of the corruption intensity is increased; the plot for \texttt{motion-blur} is displayed in \cref{fig:motion-blur}.
Here despite a general increase in error for all methods adaptive temperature scaling consistently produces lower error rates than vanilla temperature scaling (orange) and vanilla predictions (blue). 
More examples are in \cref{app:corruptions}.
\begin{table}[h]
    \centering 
\scalebox{0.7}{
	\begin{tabular}{cc|ccc|ccc}
		\toprule
		\midrule
		\textbf{Method} & \textbf{Scaling} & \textbf{Accuracy} ($\uparrow$) &  \textbf{ECE} ($\downarrow$) & \textbf{AdaECE} ($\downarrow$) & \textbf{Accuracy} ($\uparrow$) &  \textbf{ECE} ($\downarrow$) & \textbf{AdaECE} ($\downarrow$)\\
            \cmidrule(lr){1-8}
            &&\multicolumn{6}{c}{\textbf{CIFAR10-C}}\\
            \midrule
            &&\multicolumn{3}{c}{\textbf{WideResNet2810}}&\multicolumn{3}{c}{\textbf{ResNet50}}\\
            \cmidrule(lr){1-8}
            CE&None & 75.07 $\pm$ 1.46  & 15.70 $\pm$ 1.14  & 15.68 $\pm$ 1.14 & 71.45 $\pm$ 2.96  & 18.48 $\pm$ 1.70  & 18.47 $\pm$ 1.70 \\ 
            CE&Vanilla TS & 75.07 $\pm$ 1.46  & \textbf{12.19 $\pm$ 0.91}  & \textbf{12.17 $\pm$ 0.91} & 71.45 $\pm$ 2.96  & 12.72 $\pm$ 0.64  & 12.70 $\pm$ 0.63 \\ 
            CE&Adaptive TS & 75.07 $\pm$ 1.46  & \textbf{12.03 $\pm$ 1.31}  & \textbf{12.02 $\pm$ 1.31} & 71.45 $\pm$ 2.96  & \textbf{10.86 $\pm$ 1.88}  & \textbf{10.83 $\pm$ 1.87}\\
            \cmidrule(lr){1-8}
            Brier&None & 75.27 $\pm$ 0.73  & 16.21 $\pm$ 0.80  & 16.45 $\pm$ 0.78 & 74.19 $\pm$ 0.28  & 15.34 $\pm$ 0.63  & 15.34 $\pm$ 0.65 \\ 
            Brier&Vanilla TS & 75.27 $\pm$ 0.73  & 15.87 $\pm$ 0.46  & 15.86 $\pm$ 0.46 & 74.19 $\pm$ 0.28  & 14.66 $\pm$ 0.83  & 14.67 $\pm$ 0.86 \\
            Brier&Adaptive TS & 75.27 $\pm$ 0.73  & \textbf{14.84 $\pm$ 0.88}  & \textbf{14.81 $\pm$ 0.89} & 74.19 $\pm$ 0.28  & \textbf{13.39 $\pm$ 1.18}  & \textbf{13.35 $\pm$ 1.18}\\
            \cmidrule(lr){1-8}
            Ensmbls&None & 77.28  & 13.45  & 13.43 & 74.84  & 13.95  & 13.93 \\
            Ensmbls&Vanilla TS & 77.28  & 10.12  & 10.09 & 74.84  & 10.37  & 10.33 \\
            Ensmbls&Adaptive TS & 77.21  & \textbf{9.29 }  & \textbf{9.25 } & 74.80  & \textbf{9.15 }  & \textbf{9.12 }  \\
            \cmidrule(lr){1-8}
            &&\multicolumn{6}{c}{\textbf{CIFAR100-C}}\\
            \midrule
            &&\multicolumn{3}{c}{\textbf{WideResNet2810}}&\multicolumn{3}{c}{\textbf{ResNet50}}\\
            \cmidrule(lr){1-8}
            CE&None & 51.74 $\pm$ 0.39  & 18.63 $\pm$ 0.70  & 18.58 $\pm$ 0.70 & 49.67 $\pm$ 0.28  & 24.27 $\pm$ 0.89  & 24.25 $\pm$ 0.90 \\ 
            CE&Vanilla TS & 51.74 $\pm$ 0.39  & 12.28 $\pm$ 1.14  & \textbf{12.25 $\pm$ 1.14} & 49.67 $\pm$ 0.28  & \textbf{11.78 $\pm$ 0.91}  & \textbf{11.76 $\pm$ 0.91} \\ 
            CE&Adaptive TS & 51.74 $\pm$ 0.39  & \textbf{12.17} $\pm$ 0.10  & \textbf{12.15 $\pm$ 0.11} & 49.72 $\pm$ 0.29 & \textbf{11.69 $\pm$ 0.74}  & \textbf{11.67 $\pm$ 0.71}\\
            \cmidrule(lr){1-8}
            Brier&None & 50.58 $\pm$ 0.28  & 15.04 $\pm$ 1.36  & 15.02 $\pm$ 1.36 & 48.14 $\pm$ 0.83  & 13.43 $\pm$ 1.06  & 13.41 $\pm$ 1.06 \\ 
            Brier&Vanilla TS & 50.58 $\pm$ 0.28  & \textbf{9.81 $\pm$ 0.84}  & \textbf{9.81 $\pm$ 0.85} & 48.14 $\pm$ 0.83  & {10.12 $\pm$ 0.67}  & {10.10 $\pm$ 0.67} \\
            Brier&Adaptive TS & 50.58 $\pm$ 0.28  & \textbf{9.56 $\pm$ 0.82}  & \textbf{9.64 $\pm$ 0.74} & 48.62 $\pm$ 0.55  & \textbf{8.83 $\pm$ 0.48}  & \textbf{8.86 $\pm$ 0.48}\\
            \cmidrule(lr){1-8}
            Ensmbls&None & 54.61  & 14.81  & 14.78 & 52.94 & 19.12  & 19.07  \\
            Ensmbls&Vanilla TS & 54.61 & 12.66 & 12.62 & 52.94 & 11.36  & 11.33 \\
            Ensmbls&Adaptive TS & 54.61 & \textbf{12.02 }  & \textbf{12.00 } & 53.91 & \textbf{9.15 }  & \textbf{9.15 } \\
            \cmidrule(lr){1-8}
            \midrule 
            \bottomrule
            	\end{tabular}%
            }
    \caption{Corrupted calibration results. Here we can see that adaptive temperate scaling is able to improve calibration on a variety of models. Bold indicates best results, or within one standard deviation of best results.}
    \vspace{-7ex}
    \label{tab:w2810-corr-cal}
\end{table}

\subsection{Behaviour of the Temperature Prediction Module}
Can the temperature module predict high temperature in uncertain regions? If yes, then we should see a change in the temperature as we traverse the feature the space\footnote{assuming features of clean and corrupted inputs are not mapped exactly to the same point in the embedding space}.
To conduct this experiment, inspired by the analysis provided in~\cite{PintoMixMaxEnt2021,PintoRegMixup2022}, we obtain the average feature representation for each class $\bphi_k = \frac{1}{|\calX_k|}\sum_{\bx\in\calX_k}\phix$ and measure the temperature when interpolating between two classes.
i.e. we predict the temperature for the features $\{\alpha\bphi_{k^{(i)}} + (1 - \alpha)\bphi_{k^{(j)}}\}\quad\alpha\in[0, 1]$.
We plot the interpolation results in \cref{fig:interps} (Left) for the classes in CIFAR-10, where the horizontal axis represent $\alpha$ and the vertical axis represents the temperature.
For some classes we see a significant rise in the temperature as we interpolate between two classes, e.g. \texttt{automobile} and \texttt{bird}.
This highlights the temperature prediction models ability to assign a low temperature in regions that the classifier is certain about, e.g. around the mean and a higher temperature in less certain regions, e.g. heavily interpolated regions.

Interestingly, this feature is not present for all class pairs; for some, e.g. \texttt{cat} and \texttt{dog}, where the temperature remains high between classes.
We hypothesise that this is due to an interpolation between these classes being a plausible realisation of an image, unlike for \texttt{automobile} and \texttt{bird}.

We further show a histogram in \cref{fig:interps} (Right), of the temperature values for each sample in CIFAR-10 and colour code according to class.
Again we see a similar pattern where the animal based classes typically have a higher temperature than the objects, indicating that the network should be more uncertain.
This higher temperature is obtained from the VAE learning that samples in this region are often incorrect, which is where the signal comes from to increase the temperature. 



\subsection{Misclassification Rejection}
Calibrated uncertainty estimates should render that the models are able to reject samples in order to preserve the accuracy.
In this setting we report results for AURRA, which computes the area under the rejection ratio curve\cite{nadeem2009accuracy}.
We display the results in \cref{tab:aurra}, where we see that adaptive temperature scaling provides a slight improvement over normal predictions and also vanilla temperature scaling.
Furthermore, we would like to highlight that even though vanilla temperature scaling improves calibration, it does so at the expense of being able to reject samples; unlike adaptive temperature scaling which is able to provide the best of both worlds.
It is important to stress that this is a significant advantage, as we are able to provide better calibrated predictions whilst also increasing the models ability to reject samples.

\begin{table}
    \centering 
    \scalebox{0.7}{
	\begin{tabular}{c|ccc|ccc}
		\toprule
		\textbf{Methods} & \textbf{AURRA-C} ($\uparrow$) &  \textbf{AURRA-DS} ($\uparrow$) & \textbf{AURRA-E} ($\uparrow$) & \textbf{AURRA-C} ($\uparrow$) &  \textbf{AURRA-DS} ($\uparrow$) & \textbf{AURRA-E} ($\uparrow$)\\
		\cmidrule(lr){1-7}
		&\multicolumn{3}{c}{\textbf{WideResNet2810}}&\multicolumn{3}{c}{\textbf{ResNet50}}\\
		\cmidrule(lr){1-7}
		&\multicolumn{6}{c}{\textbf{CIFAR-100}}\\
		\cmidrule(lr){1-7}
		None & 93.07 $\pm$ 4.28  & 91.84 $\pm$ 5.24  & 92.95 $\pm$ 4.23 & 93.96 $\pm$ 0.15  & 92.94 $\pm$ 0.21  & 93.87 $\pm$ 0.16\\ 
        Vanilla TS & 92.97 $\pm$ 4.25  & 91.70 $\pm$ 5.40  & 92.67 $\pm$ 4.41 & 93.62 $\pm$ 0.06  & 92.68 $\pm$ 0.26  & 93.35 $\pm$ 0.11\\ 
        Adaptive TS & \textbf{93.20 $\pm$ 4.25}  & \textbf{92.00 $\pm$ 5.39}  & \textbf{92.99 $\pm$ 4.35} & \textbf{94.03 $\pm$ 0.18}  & \textbf{93.18 $\pm$ 0.19}  & \textbf{93.84 $\pm$ 0.19}\\
        \cmidrule(lr){1-7}
		&\multicolumn{6}{c}{\textbf{Tiny-ImageNet}}\\
		\cmidrule(lr){1-7}
        None & 84.21 $\pm$ 1.09  & 81.83 $\pm$ 0.64  & 83.69 $\pm$ 1.16 & 79.84 $\pm$ 2.10  & 76.71 $\pm$ 1.64  & 79.35 $\pm$ 2.25\\
        Vanilla TS & 84.05 $\pm$ 1.09  & 81.63 $\pm$ 0.64  & 83.31 $\pm$ 1.11 & 79.58 $\pm$ 2.06  & 76.27 $\pm$ 1.56  & 78.59 $\pm$ 2.07 \\
        Adaptive TS & \textbf{84.68 $\pm$ 0.18}  & \textbf{81.93 $\pm$ 0.28}  & \textbf{84.09 $\pm$ 0.20} & \textbf{81.26 $\pm$ 0.49}  & \textbf{77.60 $\pm$ 0.50}  & \textbf{80.44 $\pm$ 0.48}\\
        \bottomrule
    \end{tabular}%
        }
    \caption{AURRA scores for based on: confidence (\textbf{AURRA-C}), Demster-Schafer~\cite{sensoy2018evidential} (\textbf{AURRA-DS}) and entropy (\textbf{AURRA-E}). Unlike temperature scaling, adaptive temperature scaling does not suffer a reduction in rejection ability.}
    \label{tab:aurra}
    \vspace{-7ex}
\end{table}

\subsection{Evaluating Hardness}
Given our models ability to predict the temperature, it should naturally extract a notion of hardness, that is how difficult is it to classify.
One would expect hard samples to have a high temperature and easy ones to have a low temperature.
To conduct this experiment, we utilise the CIFAR-10.1~\cite{recht2018cifar10.1,torralba2008tinyimages} datset, which contains ``harder'', but statistically similar images to CIFAR-10; conseqently this experiment is not examining data-shift, but is instead measuring the performance on challenging samples.
We report the standard metrics: accuracy, ECE and AdaECE in \cref{tab:hardness}, where we see that adaptive temperature is able to obtain a lower calibration error than vanilla temperature scaling for both ResNet50 and WideResNet28-10 when trained using cross entropy loss.

\begin{table}
    \centering 
    \scalebox{0.8}{
    	\begin{tabular}{c|ccc|ccc}
    		\toprule
    		\textbf{Methods} & \textbf{Accuracy} ($\uparrow$) &  \textbf{ECE} ($\downarrow$) & \textbf{AdaECE} ($\downarrow$) & \textbf{Accuracy} ($\uparrow$) &  \textbf{ECE} ($\downarrow$) & \textbf{AdaECE} ($\downarrow$)\\\cmidrule(lr){1-7}
    		None & 85.86 $\pm$ 2.48  & 8.35 $\pm$ 1.58  & 8.15 $\pm$ 1.68 & 89.55 $\pm$ 0.81  & 5.66 $\pm$ 0.28  & 5.56 $\pm$ 0.32\\ 
            Vanilla TS & 85.86 $\pm$ 2.48  & 4.57 $\pm$ 0.32  & 4.24 $\pm$ 0.43 & 89.55 $\pm$ 0.81  & 3.64 $\pm$ 0.25  & 3.38 $\pm$ 0.35\\ 
            Adptive TS & 85.86 $\pm$ 2.48  & \textbf{3.67 $\pm$ 1.41}  & 3\textbf{.35 $\pm$ 1.39} & 89.55 $\pm$ 0.81  & \textbf{3.53 $\pm$ 0.22}  & \textbf{3.35 $\pm$ 0.20}\\
            \bottomrule
            	\end{tabular}
            }
    \caption{CIFAR-10.1 Results for ResNet50 and WideResNet28-10, here we see that adaptive temperature scaling is able to provide slightly improved calibration on the harder CIFAR-10.1 dataset.}\label{tab:hardness}
    \vspace{-6ex}
\end{table}

\begin{wrapfigure}{r}{0.5\textwidth}
    \centering
         \includegraphics[width=1.1\linewidth]{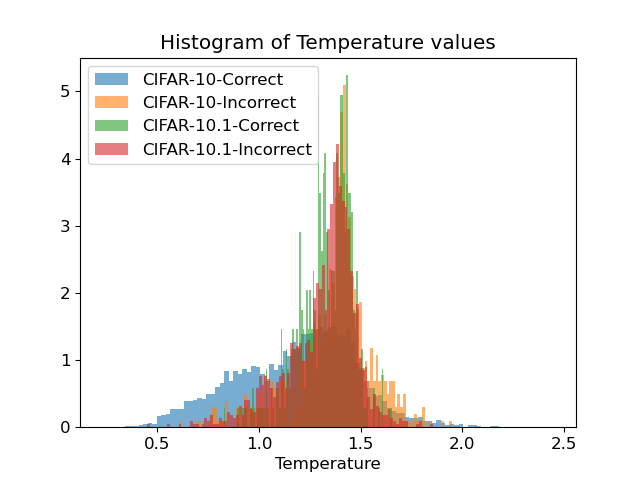}
    \caption{Histograms of temperature for correct predictions for CIFAR-10 and CIFAR-10.1 on ResNet50. Lower temperatures are typically assigned to correct (blue) samples from CIFAR-10 but higher for incorrect samples (orange). We also see that hard samples are assigned higher values, regardless of whether they are correct or not (red and green) for CIFAR-10.1.}
    \vspace{-8ex}
    \label{fig:temps_error_hist}
\end{wrapfigure}

A key hypothesis we wish to test is ``does the model assign higher temperatures to harder samples?''; harder samples should naturally contain a greater amount of uncertainty in their predictions.
Consequently, we should see higher temperature values assigned to harder samples (CIFAR-10.1) than to easier ones (CIFAR-10).
We test this hypothesis by plotting the histogram of temperature values for CIFAR-10 and CIFAR-10.1, for both correct and incorrect predictions in \cref{fig:temps_error_hist} (Right).

Here we see that generally, correct samples for CIFAR-10 (blue) are assigned a lower temperature than for CIFAR-10.1 (green), indicating that the adaptive temperature is able to recognise harder samples and assign a higher temperature increasing the uncertainty.
Furthermore, we also see that adaptive temperature predicts higher temperatures for incorrect predictions for CIFAR-10 (orange), highlighting adaptive temperatures ability to reduce the confidence of samples which are likely to be incorrect.
Interestingly, the same is not true for CIFAR-10.1, this is due to the fact that the samples from CIFAR-10.1 are by design harder, adaptive temperature predicts higher values of $T$ than for the easier CIFAR10 counterpart.

\section{Discussion}
Here we have presented a novel post-hoc method for predicting the temperature score corresponding to a given sample to make a neural network's classification confidence more calibrated. %
Given a data-point, our method is able predict how confident the classifier should be about its prediction, improving the calibration error, furthermore, adaptive temperature is also able to obtain better results under distribution shifts.
This is achieved by leveraging the latent space of a VAE, which we found to naturally encapsulate and structure the information relating to confidence appropriately.
As the model is applied post-hoc, training is very fast, requiring little computational overhead, furthermore it is very easy to implement.

\section*{Acknowledgements}
This work is supported by the UKRI grant: Turing AI Fellowship EP/W002981/1 and EPSRC/MURI grant: EP/N019474/1. We would  like to thank the Royal Academy of Engineering, FiveAI and Meta AI.  Thanks to Kemal Oksuz for spending their valuable time to provide comments on the work.

\clearpage

{\small
\bibliographystyle{splncs04}
\bibliography{arxiv}
}

\clearpage
\appendix
\section{Gradient of Network Weights}\label{app:gradients}

Consider the last layer of a Neural Network with parameters $\bw$ and the cross entropy loss $\calL : \bbR^K \rightarrow \bbR$.
The gradient of the parameters is given as $\frac{\partial\calL}{\partial\bw} = \frac{\partial\bs}{\partial\bw}\frac{\partial\sigma(\bs)}{\partial\bs}\frac{\partial\calL}{\partial\sigma(\bs)}$, where
\begin{align}
    \frac{\partial\calL}{\partial\sigma(s_k)} &= -\frac{q_k}{\sigma(s_k)}\\
    \frac{\partial\sigma(s_k)}{\partial s_j} &= \sigma(s_k)(\delta_{jk} - \sigma(s_j)).
\end{align}
the gradient for the last layers is thus given as
\begin{align}
    \frac{\partial\calL}{\partial\bw} = \frac{\partial\bs}{\partial\bw}(\sigma(\bs) - \bq),
\end{align}
where $\sigma(\bs) - \bq = \{\sigma(s_j) - q_j : j \in \{1\dots K\}$.

\section{Predictions are unaffected by temperature}\label{app:temp}
In neural network classification problems, the parameters of the Categorical distribution are obtained through the Softmax operator
\begin{align*}
    \sigma(\bs) = \frac{\exp{(\frac{\bs}{T})}}{\sum_i \exp{(\frac{s_i}{T})}},
\end{align*}
with the predicted class given as $\tilde{k} = \argmax_k\sigma(\bs_k)$. 
The temperature value has no effect on the resulting prediction
\begin{align}
    \argmax_k\sigma(\bs_k) &= \argmax_k\frac{\bs_k}{T} \\
                           &= \argmax_k\bs_k. 
\end{align}
Hence, the value of $T$ does not affect the class prediction.

\section{Gradient of Temperature}\label{app:temp_grad}
The gradient of the loss w.r.t the temperature is $\frac{\partial\calL}{\partial T} = \frac{\partial\bp}{\partial T}\frac{\partial\calL}{\bp}^T$.
The gradient of the individual class probabilities from the softmax output is
\begin{align*}
    \frac{\partial \bp_k}{\partial T} &= \frac{\sum_i\exp{(\frac{\bs_i}{T})}\frac{\partial}{\partial T}\exp{(\frac{\bs_k}{T})}}{\big(\sum_i\exp{(\frac{\bs_i}{T})}\big)^2} - 
                                      \frac{\exp{(\frac{\bs_k}{T})}\sum_k\frac{\partial}{\partial T}\exp{(\frac{\bs_k}{T})}}{\big(\sum_i\exp{(\frac{\bs_i}{T})}\big)^2} \\
                                    &= \frac{\sigma(\bs_k)}{T^2}\Big(\sum_{i\text{\textbackslash} k}\bs_i\exp{(\frac{\bs_i}{T})} - \bs_k\exp{(\frac{\bs_i}{T})}\Big).
\end{align*}
Given that $\frac{\partial\calL}{\partial\sigma(s_k)} = -\frac{q_k}{\sigma(s_k)}$, using the chain rule we have
\begin{align}
    \frac{\partial\calL}{\partial T} = \sum_k \frac{q_k}{T^2}\bigg(s_k\sum_{i\text{\textbackslash}k}\exp\Big(\frac{s_i}{T}\Big) - \sum_{j\text{\textbackslash}k}s_j\exp\Big(\frac{s_j}{T}\Big)\bigg),
\end{align}
thus concluding the proof.

\section{Corruptions}\label{app:corruptions}
\begin{figure}
    \begin{tabular}{ccc}
        \includegraphics[width=0.32\linewidth]{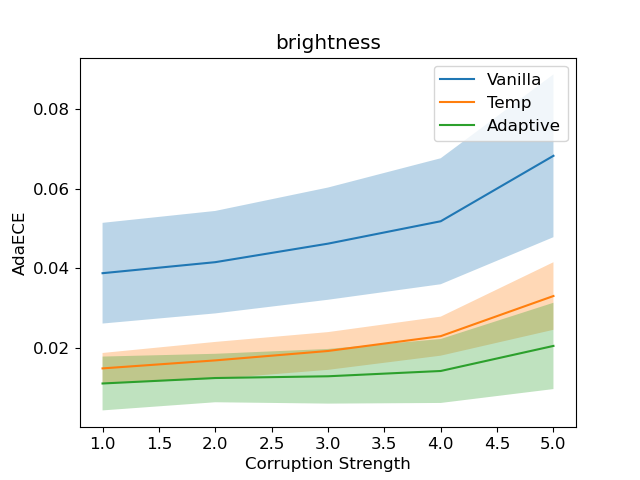} & 
        \includegraphics[width=0.32\linewidth]{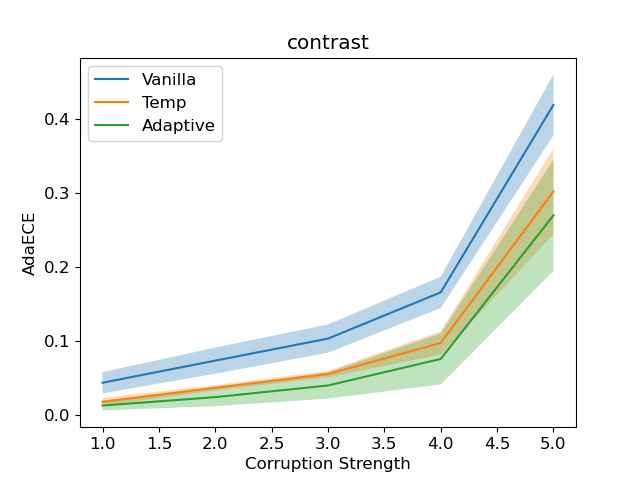} & 
        \includegraphics[width=0.32\linewidth]{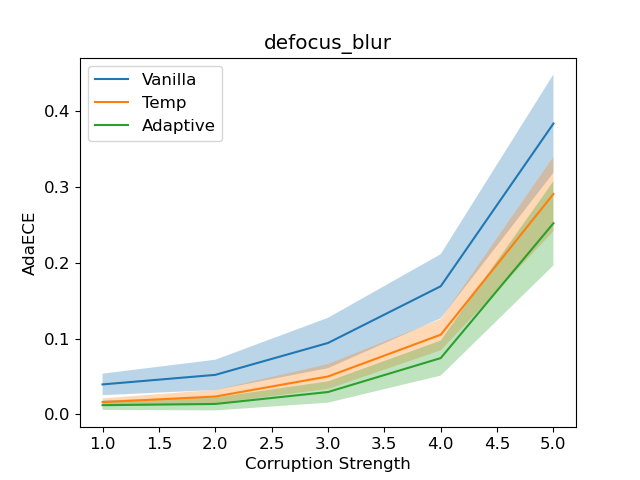} \\
        \includegraphics[width=0.32\linewidth]{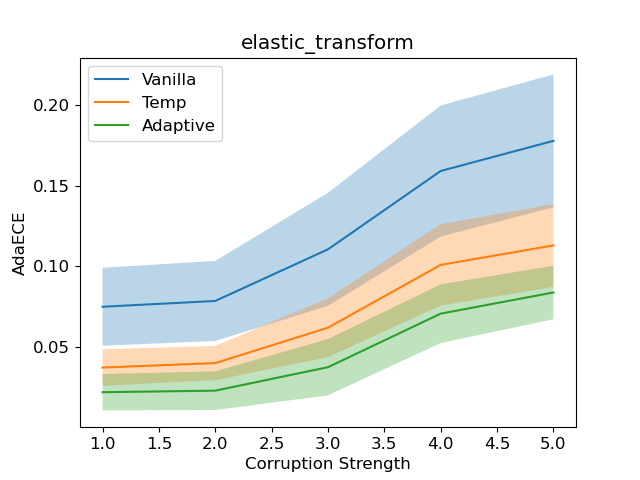} & 
        \includegraphics[width=0.32\linewidth]{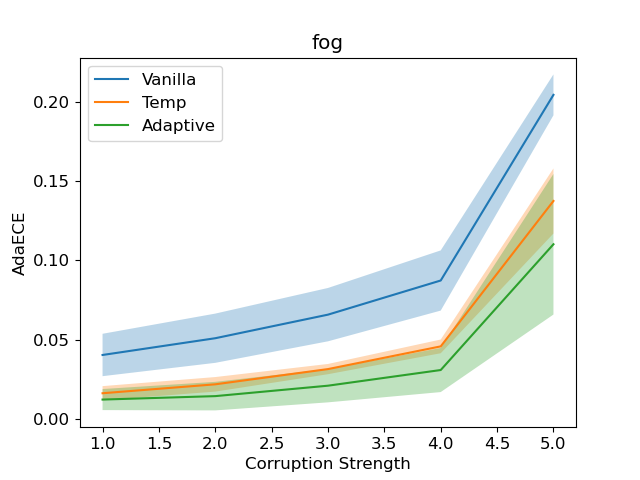} & 
        \includegraphics[width=0.32\linewidth]{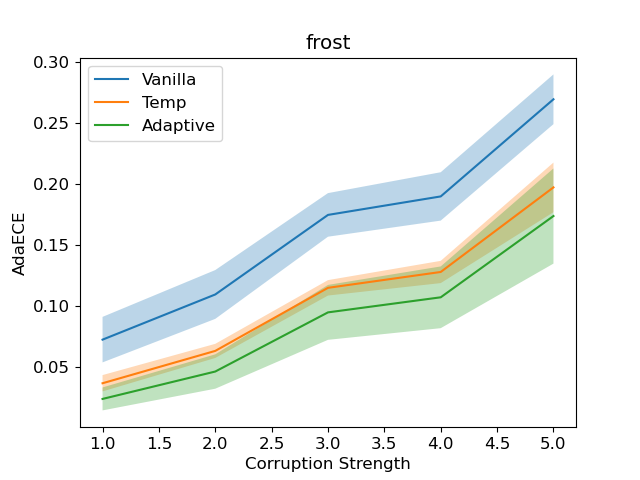} \\
        \includegraphics[width=0.32\linewidth]{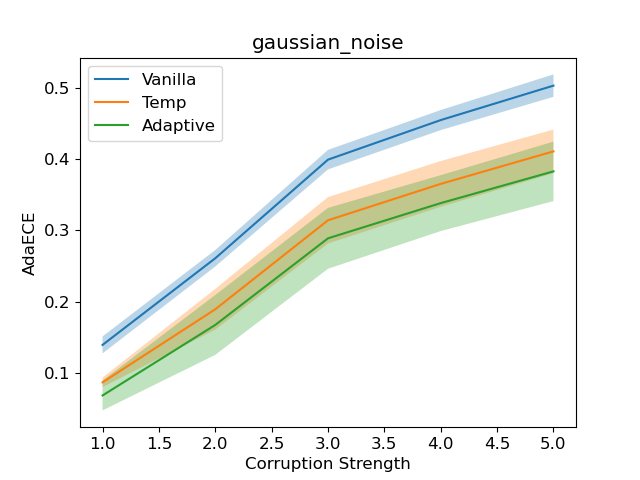} & 
        \includegraphics[width=0.32\linewidth]{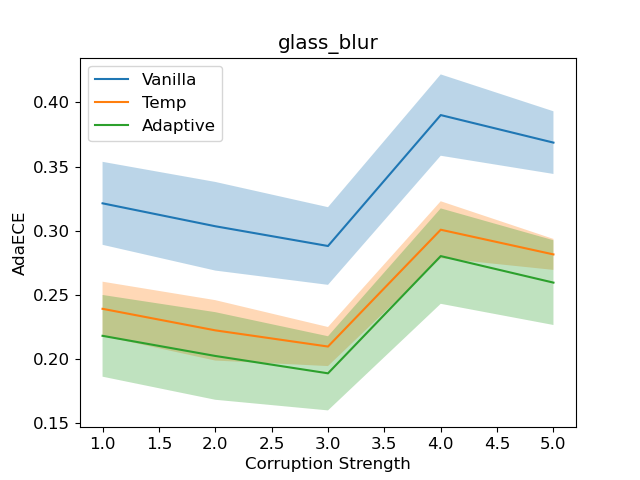} & 
        \includegraphics[width=0.32\linewidth]{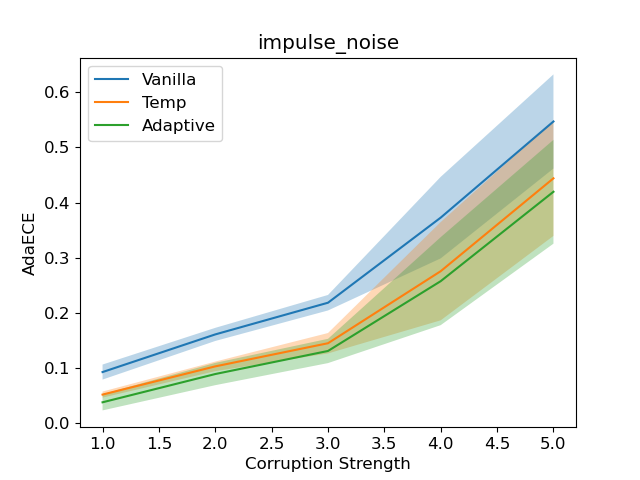} \\
        \includegraphics[width=0.32\linewidth]{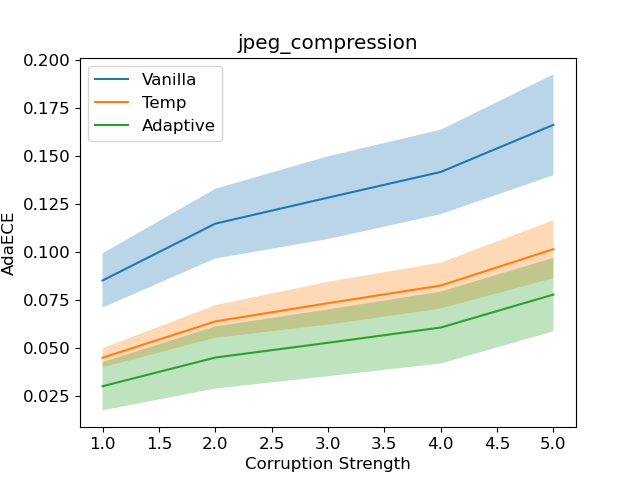} & 
        \includegraphics[width=0.32\linewidth]{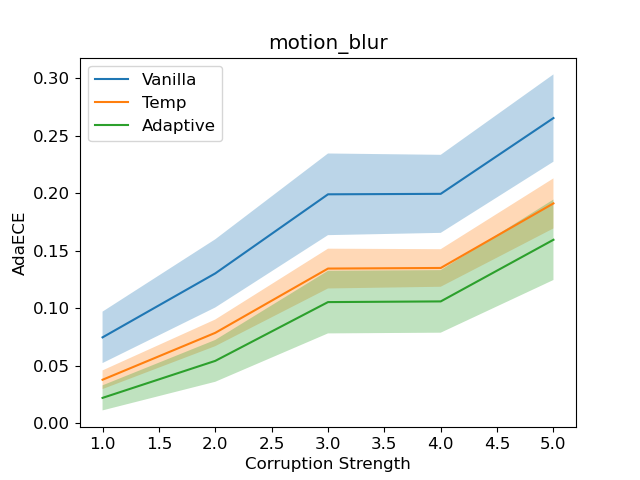} & 
        \includegraphics[width=0.32\linewidth]{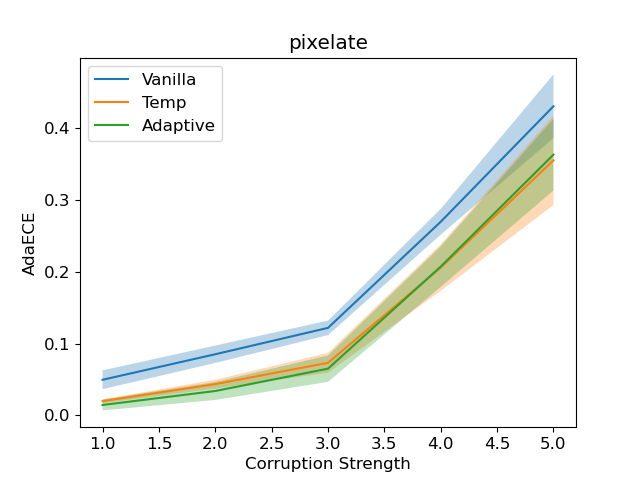} \\
        \includegraphics[width=0.32\linewidth]{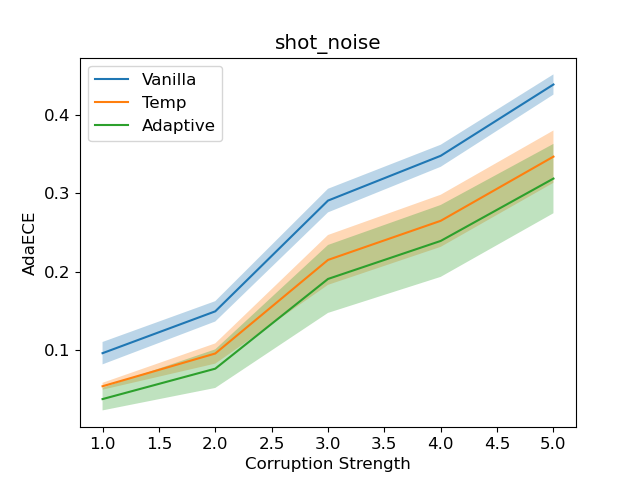} & 
        \includegraphics[width=0.32\linewidth]{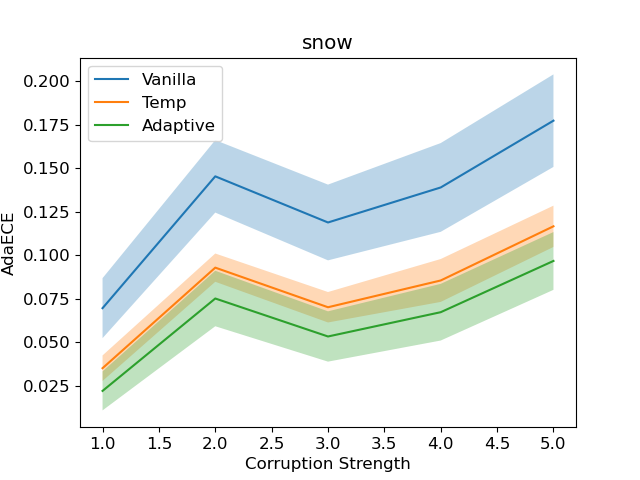} & 
        \includegraphics[width=0.32\linewidth]{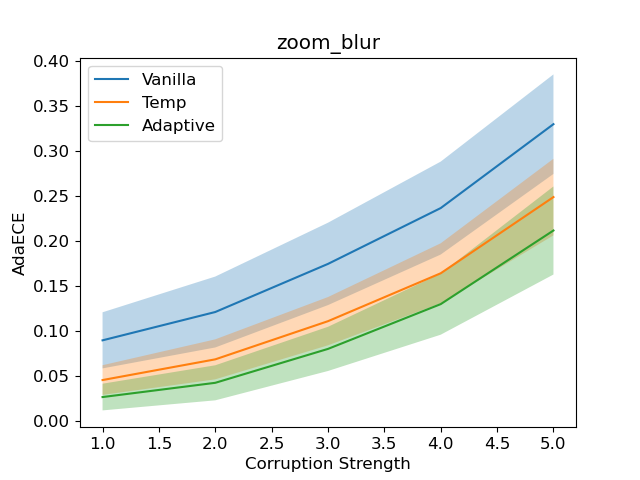}
    \end{tabular}
    \caption{Additional plots for how AdaECE varies with corruption strength for CIFAR10-C.}
    \label{fig:more_corrs}
\end{figure}

We display additional plots for how AdaECE varies with corruption strength for CIFAR10-C in~\cref{fig:more_corrs}, where we can see that adaptive temperature scaling consistently obtains better results than vanilla temperature scaling.

\section{Temperature Values}
We display the average temperature values in \cref{tab:temp_vals}, here we see that adaptive temperature obtains a similar average temperature to vanilla temperature scaling.

\begin{table}
	\centering 
	\scalebox{0.8}{
		\begin{tabular}{c|c|cc}
			\toprule
			\textbf{Network} & \textbf{Dataset}  &  \multicolumn{2}{c}{\textbf{Avg. Temp}}\\
			&& \emph{Vanilla} & \emph{Adaptive} \\\cmidrule(lr){1-4}
			\multirow{3}{*}{ResNet50}& CIFAR-10 & 1.484 $\pm$ 0.123 & 1.506 $\pm$ 0.296  \\
			 & CIFAR-100  & 1.398 $\pm$ 0.034 & 1.313 $\pm$ 0.076  \\
			& TinyImageNet & 1.296 $\pm$ 0.234 & 1.131 $\pm$ 0.179 \\\cmidrule(lr){1-4}
			\multirow{3}{*}{WideResNet2810}& CIFAR-10 & 1.310 $\pm$ 0.035 & 1.294 $\pm$ 0.072  \\
			& CIFAR-100  & 1.220 $\pm$ 0.010 & 1.134 9  \\
			& TinyImageNet & 1.174 $\pm$ 0.084 & 1.017 $\pm$ 0.117 \\
			\bottomrule
		\end{tabular}
	}
	\caption{Average temperature values for neural different models.}\label{tab:temp_vals}
	\vspace{-3ex}
\end{table}

\section{Training Details}
We followed standard training protocols when training the neural networks.
Models trained on CIFAR-10/CIFAR-100 required 350 epochs, with an initial learning rate of 0.1, the learning rate was decreased by a factor of 10 at the milestones 150 and 250 epochs.
Models trained on TinyImageNet required 100 epochs, with an initial learning rate of 0.1, the learning rate was decreased by a factor of 10 at the milestones 43 and 72 epochs.

\end{document}

%% file: math.tex
\newcommand{\bx}{\mathbf{x}}
\newcommand{\bz}{\mathbf{z}}

\newcommand{\by}{\mathbf{y}}
\newcommand{\bs}{\mathbf{s}}

\newcommand{\bbE}{\mathbb{E}}

\newcommand{\bbR}{\mathbb{R}}

\newcommand{\calL}{\mathcal{L}}

\newcommand{\qzx}{q_\phi(\bz\mid\bx)}

\newcommand{\pzy}{p_{\lambda_k}(\bz\mid\by)}

\DeclareMathOperator*{\argmax}{arg\,max}

\newcommand{\bw}{\boldsymbol{w}}
\newcommand{\bp}{\boldsymbol{p}}
\newcommand{\bq}{\boldsymbol{q}}
\newcommand{\bphi}{\boldsymbol{\phi}}

\newcommand{\phix}{\Phi(\bx)}

\newcommand{\softmax}[1]{\text{softmax}(#1)}

\newcommand{\elbo}[1]{\mathbb{ELBO}[#1]}

\newcommand{\calN}{\mathcal{N}}
\newcommand{\calX}{\mathcal{X}}
\newcommand{\calP}{\mathcal{P}}
\newcommand{\calY}{\mathcal{Y}}